\documentclass{article} 
\usepackage{iclr2024_conference,times}
\iclrfinalcopy

\usepackage{amsmath,amsfonts,bm}

\def\eqref#1{equation~\ref{#1}}

\def\1{\bm{1}}

\DeclareMathAlphabet{\mathsfit}{\encodingdefault}{\sfdefault}{m}{sl}
\SetMathAlphabet{\mathsfit}{bold}{\encodingdefault}{\sfdefault}{bx}{n}

\usepackage{url}
\usepackage[utf8]{inputenc} 
\usepackage[T1]{fontenc}    
\usepackage[colorlinks,citecolor=blue]{hyperref}
\usepackage{url}            
\usepackage{booktabs}       
\usepackage{amsfonts}       
\usepackage{nicefrac}       
\usepackage{microtype}      
\usepackage{tabularx}
\usepackage{tcolorbox}
\usepackage{amsthm, amssymb, amsmath}
\usepackage{verbatim, float}
\usepackage{mathrsfs}
\usepackage{graphics}
\usepackage{inconsolata}
\usepackage{subfig}
\usepackage[usenames,dvipsnames]{pstricks}
\usepackage{multirow}
\usepackage{url}
\usepackage{array}
\usepackage{tabu}
\usepackage{epsfig}
\usepackage{parallel}
\usepackage{graphicx}
\usepackage{diagbox}
\usepackage{mathtools}
\usepackage[absolute,overlay]{textpos}
\usepackage{wrapfig}
\usepackage[linesnumbered,vlined,ruled,commentsnumbered]{algorithm2e}\usepackage{listings}
\usepackage{xcolor}
\usepackage{enumitem}

\definecolor{codegreen}{rgb}{0,0.6,0}
\definecolor{codegray}{rgb}{0.5,0.5,0.5}
\definecolor{codepurple}{rgb}{0.58,0,0.82}
\definecolor{backcolour}{rgb}{0.95,0.95,0.95}

\lstdefinestyle{mystyle}{
  backgroundcolor=\color{backcolour},
  commentstyle=\color{codegreen},
  keywordstyle=\color{magenta},
  numberstyle=\tiny\color{codegray},
  stringstyle=\color{codepurple},
  basicstyle=\ttfamily\footnotesize,
  breakatwhitespace=false,
  breaklines=true,
  captionpos=t,
  keepspaces=true,
  numbers=left,
  numbersep=5pt,
  showspaces=false,
  showstringspaces=false,
  showtabs=false,
  tabsize=2
}

\newcommand{\ours}{\texttt{TPD}\xspace}

\title{
  \ours: Enhancing Student Language Model Reasoning via Principle Discovery and Guidance
}

\author{Haorui Wang$^{1}$, Rongzhi Zhang$^1$, Yinghao Li$^1$, Lingkai Kong$^1$, Yuchen Zhuang$^1$,\\ \textbf{Xiusi Chen$^2$, Chao Zhang$^1$} \\
  $^1$College of Computing, Georgia Institute of Technology \\
  $^2$Department of Computer Science, University of California, Los Angeles \\
  \texttt{\{hwang984,rongzhi.zhang,yinghaoli,lkkong,yczhuang,chaozhang\}@gatech.edu},\\\,\texttt{xchen@cs.ucla.edu}}

\begin{document}

\maketitle

\begin{abstract}

  Large Language Models (LLMs) have recently showcased remarkable reasoning abilities. However, larger models often surpass their smaller counterparts in reasoning tasks, posing the challenge of effectively transferring these capabilities from larger models. Existing approaches heavily rely on extensive fine-tuning data or continuous interactions with a superior teacher LLM during inference.
  We introduce a principle-based teacher-student framework called ``Teaching via Principle Discovery'' (\ours) to address these limitations. Inspired by human learning mechanisms, \ours mimics the interaction between a teacher and a student using a principle-based approach. The teacher LLM generates problem-solving instructions and corrective principles based on the student LLM's errors.
  These principles guide the refinement of instructions and the selection of instructive examples from a validation set. This enables the student model to learn from both the teacher's guidance and its own mistakes. Once the student model begins making inferences, \ours requires no further intervention from the teacher LLM or humans.
  Through extensive experiments across eight reasoning tasks, we demonstrate the effectiveness of \ours. Compared to standard chain-of-thought prompting, \ours significantly improves the student model's performance, achieving  6.2\% improvement on average.

\end{abstract}
\section{Introduction}

Recent studies show that large language models (LLMs) can achieve impressive performance in various reasoning tasks, such as analogical \citep{webb2023emergent}, arithmetic \citep{imani2023mathprompter}, symbolic \citep{pan2023logic}, and commonsense reasoning \citep{wei2022chain, bang2023multitask}.
However, a noticeable performance gap can often be observed between stronger LLMs such as GPT-4 and weaker LLMs such as GPT-3.5-turbo \citep{espejel2023gpt, DBLP:journals/corr/abs-2303-08774, cai2023large}.
This disparity arises from factors such as training data size, model capacity, and the methods by which LLMs learn and encode world knowledge. While stronger LLMs exhibit superior performance, their practical application is hindered by the high costs associated with training and inference. For instance, as of this writing, the cost of using GPT-4 is over ten times higher compared to GPT-3.5-turbo. This raises the question: how can we effectively transfer the advanced reasoning capabilities of stronger LLMs to weaker ones?

Several approaches have been proposed to address this challenge.
For example, some studies \citep{DBLP:conf/acl/RajaniMXS19,DBLP:conf/acl/HoSY23} curate datasets for specific downstream tasks using stronger LLMs and then fine-tune weaker LLMs on these datasets to instill the necessary knowledge.
However, this fine-tuning process is time-consuming, and the resulting task-specific weaker LLMs lack generalizability to other tasks.
Other methods \citep{wang2023learn} involve using an assistant language model to provide guidelines and analysis for the student model, but this approach requires constant involvement of the assistant model, which can be costly.
To reduce teacher model involvement, \citet{saha2023can} request teacher intervention only when the student model exhibits low confidence, where the confidence is computed from the token probability of the answer.
Such a measure may not accurately reflect the student model's true confidence, as the token probability is influenced by the preceding output, specifically the chain-of-thought (CoT) explanations.

\begin{wrapfigure}{r}{0.5\textwidth}
  \centering
  \includegraphics[width=0.48\textwidth]{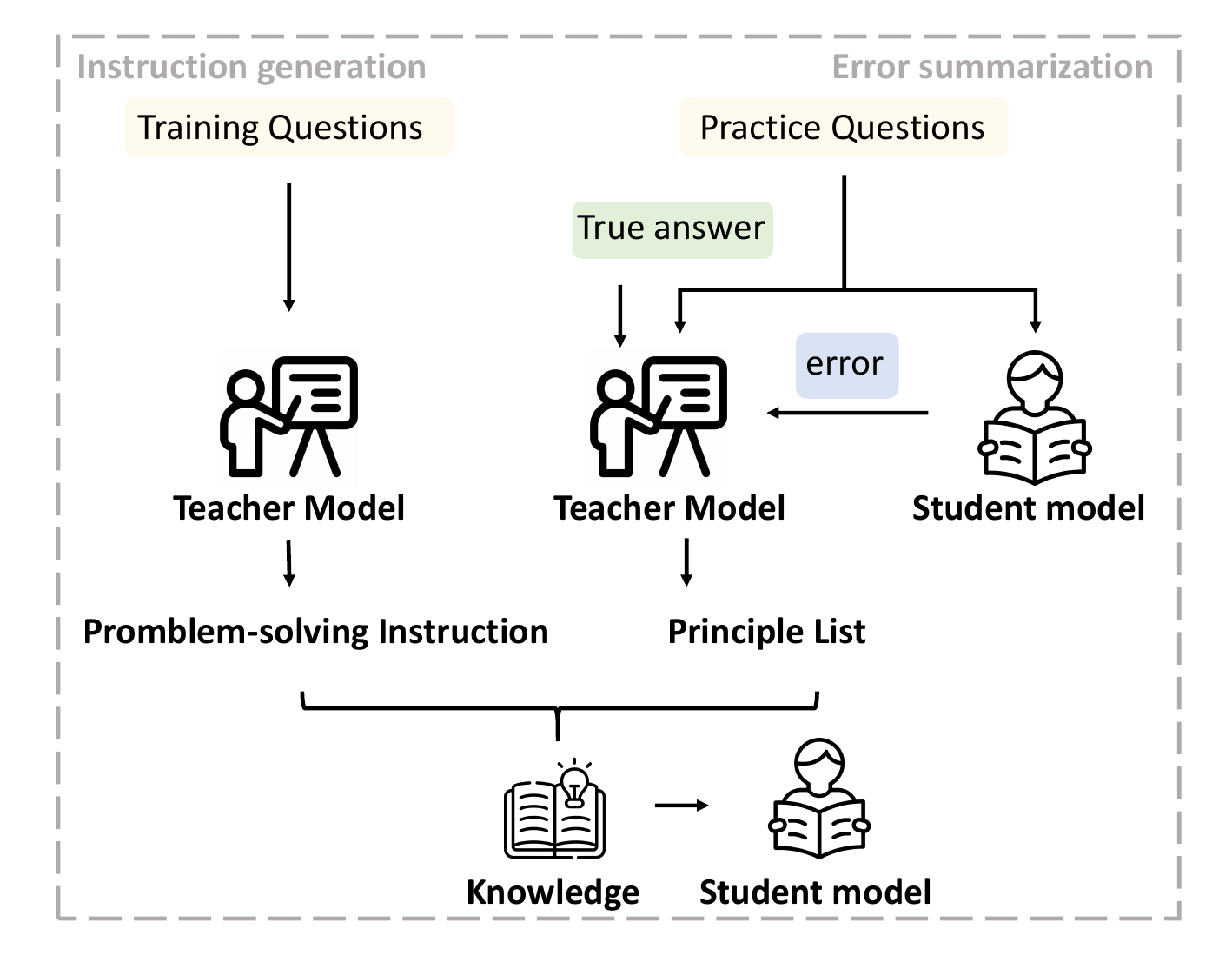}
  \caption{
    Illustration of \ours.
    To generate corrective principles, the teacher model first generates problem-solving instructions and then summarizes principles based on errors made by the student model on validation questions.
    During principle exploitation, the problem-solving instruction and examples that illustrate the principles are combined into the prompt to guide student learning.
  }
  \label{fig:intro_overview}
\end{wrapfigure}

To address these limitations, we introduce Teaching via Principle Discovery (\ours), a principle-based teaching framework that minimizes teacher model involvement, thereby optimizing resource allocation and efficiency.
\ours draws inspiration from instructional strategies observed in natural human teaching and learning processes, as outlined in studies such as \citep{rosenshine2012principles, henderson2009quiz, metcalfe2017learning}. The framework follows a structured "Demonstrate-Practice-Review" process: 1) \emph{Demonstrate}: The teacher model introduces a problem type and demonstrates how to solve it. 2) \emph{Practice}: The student model engages with practice questions. 3) \emph{Review}: The teacher model reviews the student's responses to identify common errors. This review process enables the teacher model to extract fine-grained \emph{corrective principles} that guide the student in rectifying errors and improving future problem-solving. Inspired by this process, \ours comprises two stages: \emph{principle generation} and \emph{principle exploitation}. In the principle generation stage, the teacher model generates problem-solving instructions and summarizes principles based on errors made by the student model on validation questions. In the principle exploitation stage, the teacher constructs instructive examples that illustrate the principles and injects these instructions into the prompt to guide student learning.

We validate our method on eight reasoning tasks covering symbolic and arithmetic reasoning.
Through principle generation and exploitation, our model significantly improves the performance of the student model on the test set without teacher model intervention during inference.
Specifically, our model achieves an absolute gain of up to 19\% accuracy
 compared to  chain-of-thought prompting.
Additionally, we explore effective methods for injecting generated principles into the student model and find that selecting new examples from practice questions outperforms direct injection of the principle list or the critique-revise method.

Our key contributions are summarized as follows:
\begin{itemize}[nosep,leftmargin=*]
  \item

        We introduce \ours, a teacher-student framework that enhances the effectiveness of teacher models in educating student models.
  \item

        By leveraging problem-solving instruction and error summarization, the student model learns from the teacher's guidance and its own past mistakes.

  \item

        \ours reduces teacher model involvement during testing, enabling the student model to operate independently in offline scenarios.

\end{itemize}

\section{Related work}

\paragraph{Teacher-student framework}
The teacher-student framework aims to transfer knowledge from a larger teacher model to a smaller student model \citep{gou2021knowledge}. Traditional approaches have employed fully supervised fine-tuning methods to achieve this goal \citep{magister2022teaching, shridhar2022distilling}. For instance, ~\citet{DBLP:conf/acl/RajaniMXS19} build the Common Sense Explanations (CoS-E) dataset by generating explanations using GPT~\citep{radford2018improving} to finetune BERT models, improving their commonsense reasoning abilities.
\citet{hendrycks2021measuring} take a similar approach by finetuning models on datasets featuring step-by-step mathematical problem solutions. \citet{DBLP:conf/acl/HoSY23} generate CoT reasoning steps by GPT-3 \citep{brown2020language} and finetune several relatively small models with these generated data. However, both the fine-tuning process and collecting task-specific fine-tuning data are time-consuming and challenging. Moreover, these fine-tuned models are often dataset-specific, limiting their applicability across diverse tasks. To overcome these limitations, there are several prompting-based teacher-student frameworks recently \citep{pruthi2022evaluating, saha2023can, yu2023characterizing}. In these frameworks, the teacher model is tasked with offering demonstrations or explanations to the student model. However, these approaches require the intervention of a teacher model on the test set. Hence, they are unsuitable for offline scenarios, where the student model must solve problems independently, without assistance from the teacher model.

\paragraph{Eliciting LLM's reasoning ability through prompting}
Emergent abilities bring a strong few-shot learning ability for LLMs across various datasets via in-context learning \citep{DBLP:journals/tmlr/WeiTBRZBYBZMCHVLDF22, wei2022chain}. Many prompting-based methods are proposed to elicit the reasoning abilities of the LLMs by injecting knowledge into prompts. For instance, ``chain of thought'' (CoT) ~\citep{wei2022chain} and its variants \citep{kojima2022large, DBLP:journals/corr/abs-2301-00303} provide a few human written examples or instructions to LLMs. ~\citet{DBLP:conf/emnlp/MadaanZ0YN22, DBLP:conf/icml/GaoMZ00YCN23, DBLP:journals/corr/abs-2211-12588} find that using LLMs to generate codes for reasoning tasks and then utilizing the codes by program interpreters to solve the questions can achieve better performance. Another prompting method is the problem decomposition method ~\citep{zhou2022least, drozdov2022compositional, dua2022successive}, which asks LLMs to decompose the tasks into several subquestions and answer them individually. Self-refinement methods ~\citep{madaan2023self} ask LLMs to refine their original answers iteratively but will rely on the handwritten few shot examples. These prompting methods are effective in eliciting the reasoning ability of the LLMs. However, they are not suitable for direct adaptation within the teacher-student framework, as they do not involve knowledge transfer from a teacher model.

\paragraph{Principle discovery and exploitation}
Our method uses advanced models to generate principles to teach weaker agents to do reasoning tasks, which is also related to principle/rule discovery. Rule discovery is a popular technique in machine learning and data mining \citep{tweney1980strategies, furnkranz2015brief, das1998rule}. 
In language models, rule discovery is often utilized in learning from feedback frameworks \citep{zhang2022prboost, pan2023automatically}. In \citep{zhu2023large}, rules are generated by language models in high temperatures and are subsequently selected through a self-verification process. \citet{yang2023failures} propose a framework where language models could generate rules by learning from previous mistakes as data streams in.. However, the rules produced in these methods tend to be concrete facts, with the framework functioning as a retrieval system. In contrast, a rule is a formal, broad statement that applies to an indefinitely large set of objects. As such, relying solely on text similarity to uncover necessary rules is impractical. Unlike rules, principles are more abstract and open to interpretation, offering high-level guidance without strict formatting. For instance, \citet{bai2022constitutional} predefine principles about helpfulness and harmfulness by human experts, then ask the language model to evaluate the generation results to make RLAIF.

\section{Method}
\begin{figure}[t!]
\centering
    \includegraphics[width=\textwidth]{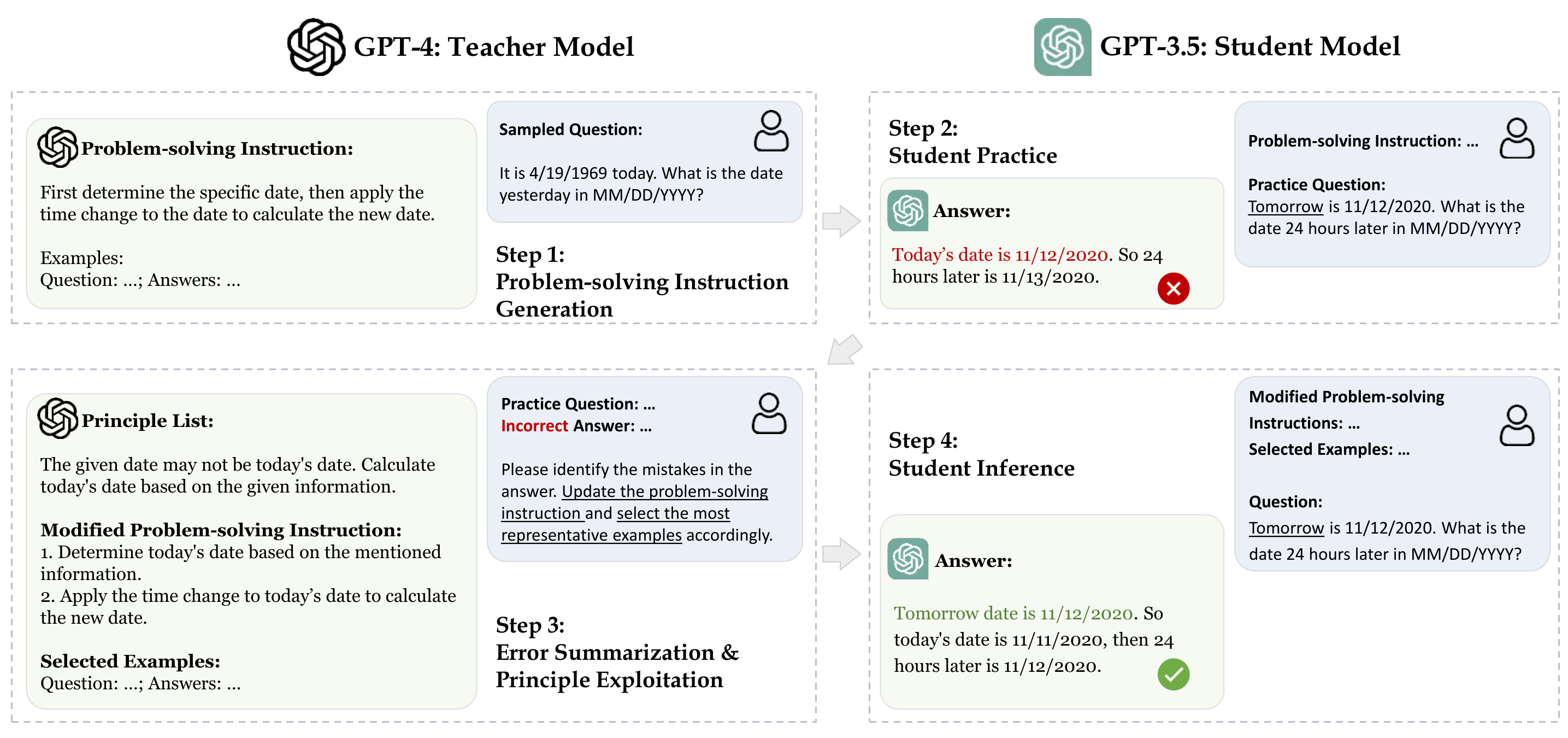}
    \caption{\ours pipeline. It contains two stages: principle generation and principle exploitation. In principle generation, the student model generates answers according to the problem-solving instructions from the teacher model. Then, the teacher provides a list of principles based on student's practice errors. In principle exploitation, the teacher model refines the instruction and chooses representative examples, which are used by the student for inference.
    }
    \label{fig:pipeline}
\end{figure}

\subsection{Overview}
In \ours, we consider a teacher model and a student model. The student is responsible for solving reasoning problems based on the teacher model's guidance. The primary goal of our framework is to let the teacher model effectively teach the student model how to solve these tasks. We denote the model response as $f(m, p)$ given the LLM $m$ and the input $p$.

As shown in Fig.~\ref{fig:pipeline}, the framework is divided into two stages: \textbf{principle generation} and \textbf{principle exploitation}.
During the principle generation stage, the teacher model produces a problem-solving instruction based on sampled questions and identifies a principle list $\mathbb{P}$ from the errors made by the student model. In the principle exploitation stage, the principle list $\mathbb{P}$ is injected into the student model to improve its reasoning capabilities in downstream tasks.

\subsection{Principle generation}


In this stage, we employ the teacher model to provide a high-level guideline to the student model and help the student model identify its own common errors from a few practice questions. This principle generation stage comprises two sub-stages: 1. Problem-solving instruction generation and initial practice and 2. Error summarization. This two-stage process mirrors the traditional classroom learning model, where teachers provide instructions, students practice, and teachers offer feedback based on observed errors.


\subsubsection{Problem-solving instruction generation and initial practice}

\paragraph{What is problem-solving instruction?}
The problem-solving instruction $I$ consists of a problem-solving method augmented with a few examples. To build the instruction, the teacher model reviews a set of questions sampled from the training set, identifying what type of questions the student model needs to solve. Then, the teacher model generates a problem-solving method as an initial instruction in natural language to solve this type of problem. To enhance the instruction, it is further enriched with examples that demonstrate the application of the proposed method. These examples are derived from the same questions initially sampled. We provide an example below for a better understanding.
\begin{tcolorbox}[title=Problem-solving Instruction:]\small

To solve the problem where you are asked to take the last letters of each word in a given string and concatenate them, you can follow these steps:\\
1. Identify and list each word in the string.\\
2. Locate the last letter of each word.\\
3. Concatenate these letters to form a new string. Provide the new string as the answer.

\textbf{Examples:}
Question: Take the last letters of each word in "Jeremiah Kelley Josue Veronica" and concatenate them. \\
Identify the words: Jeremiah, Kelley, Josue, Veronica. \\
Last letters: h, y, e, a. \\
Concatenate: "hyea". \\
Answer: hyea.
\end{tcolorbox}

\paragraph{Why problem-solving instruction?}

Previous works \citep{saha2023can, wang2023learn} ask the teacher model to offer guidance for each test question, rendering the teaching framework impractical for offline scenarios.
To tackle this issue, we ask the teacher model to provide high-level problem-solving instruction based on sampled questions and internal knowledge. The problem-solving instruction is adaptable, making it suitable for offline use as it generalizes across similar types of questions.
Moreover, since many prompting methods \citep{wei2022chain, gao2023pal,zhou2022least} have shown that LLMs have a strong capability to learn from examples, the problem-solving instruction includes a few examples helping the student model understand and imitate the instruction. This process also follows the ``Case Teaching Method'' paradigm \citep{herreid2005using}, where several cases are provided by the teacher models, and the student model can imitate and learn from the examples. 
The problem-solving instruction reflects the key principles to solve the type of problems, as it provides fundamental guidance and rules for addressing specific types of problems.

\paragraph{Initial practice for the student model}
With the problem-solving instruction, the student model generates answers for the questions in the validation set. This step lets the student model practice, allowing the teacher model to assess its comprehension of the problem-solving instruction and its ability to effectively apply it.

\subsubsection{Error Summarization}
\paragraph{Data filtering} After the student practices over the validation set, we evaluate its responses to build an error set $\mathbb{E} = \{e_1, e_2, ..., e_n\}$, where each element $e_i$ in this set represents a pair consisting of a question and the corresponding incorrect answer made by the student model. Although the teacher model generally performs well on various tasks, it can make mistakes. Hence, we need to assess the teacher model's ability to accurately identify the incorrect answers in $\mathbb{E}$. Specifically, we check if the teacher model agrees with the student's incorrect answers, removing each $e_i$ that the teacher model can not identify to build a feasible error set $\mathbb{\hat{E}}$. Due to the context length of the LLMs, we then develop an iterative framework to summarize principles from each $e_i$ iteratively.

\begin{wrapfigure}{l}{0.48\textwidth}
\centering
\begin{algorithm}[H]
\label{error_algo}
    \SetAlgoLined 
	\caption{Error summarization}
	\KwIn{$T$: the teacher model; $\mathbb{\hat{E}}$: the feasible error set}
	\KwOut{$\mathbb{P}$: the principle list}
        Sample a subset $\mathbb{N}_s$ from $\mathbb{\hat{E}}$\;
        $\mathbb{N}_r \leftarrow \mathbb{\hat{E}} \backslash \mathbb{N}_s$\;
        $\mathbb{P} \leftarrow f_{\rm summarize}(T, \mathbb{N}_s)$\;
        \For{$i \in \{1,2,3,\dots ,|\mathbb{N}_r|\}$}{
            \If{$f_{\rm evaluate}(T, \mathbb{P}, \mathbb{N}_r(i)) = \text{False}$ }{
                $p \leftarrow f_{\rm summarize}(T, \mathbb{N}_r(i))$\;
                $\mathbb{P} \leftarrow \mathbb{P} \cup p$\;
            }
        }
	return $\mathbb{P}$
\end{algorithm}
\vspace{-0.3cm}
\end{wrapfigure}

\paragraph{What is error summarization?} As illustrated in Algorithm~\ref{error_algo}, the teacher model initially derives principles $\mathbb{P}$ from a subset $\mathbb{N}_s$ of the feasible error set $\mathbb{\hat{E}}$.
Then, we present the remaining set $\mathbb{N}_r$ to the teacher model sequentially, prompting it to determine whether the established principle list $\mathbb{P}$ can rectify the issue. If the existing principle list $\mathbb{P}$ can not address the presented error $\mathbb{N}_r(i)$, the teacher model will formulate a new principle $p$ for it. The iterative process stops when all the errors in $\mathbb{N}_r(i)$ are checked.
Then human reviewers will step in to assess the validity of the principle list $\mathbb{P}$ created by the teacher model. The reviewers will simply delete any that are found to be erroneous or confusing, ensuring the reliability and clarity of the final principle list.

\paragraph{Why error summarization?}
Making mistakes is a natural part of the learning process and can be a powerful tool for growth and understanding \citep{cyr2018learning}. Even though students can solve simple tasks using the meta instruction alone (e.g., executing a single technique), they may have difficulty applying the same knowledge and skills in complex situations \citep{klein2007genetically}. Therefore, it is also important to let students learn from errors, which can help them identify error patterns and avoid similar mistakes in future practice \citep{metcalfe2017learning}. Learning from mistakes has been explored in \citep{wang2023learn, yang2023failures}, however, they only record the errors in detail and do not summarize these mistakes to high-level principles. The high-level principles are more generalizable and can be easily applied to offline scenarios.

\subsection{Principle exploitation}
The second stage lets the student model use the generated principle list $\mathbb{P}$. The student model's role is to utilize the principle lists to solve various instances of the task. There are multiple ways to utilize the principle list, which are explored in Sec~\ref{use_principle}. Compared with directly injecting the principle list into the prompt and the critique-revise method, we found that a more effective approach is to utilize the principle list $\mathbb{P}$ to curate new examples from the validation set. Specifically, we define the violation score, which is the number of violations against the principles in the list $\mathbb{P}$. The teacher model evaluates each error in $\mathbb{\hat{E}}$, using violation scores to rank them. Then the teacher model generates correct examples for the error with the highest violation scores. Besides, the teacher model also revises the problem-solving instruction based on $\mathbb{P}$. The overall instruction is a combination of the revised problem-solving instruction and the selected informative examples. For better illustration, we present the overall instruction template below.

\begin{tcolorbox}[title=Overall Instruction Prompt]\small
    Revised Problem-solving Instruction: \{\emph{method}\} + \{\emph{original examples}\} \\
    New Selected Examples: \{\emph{question}\} + \{\emph{answer}\}
\end{tcolorbox}

The principle generation stage only needs to be performed once for each type of task. Then the teacher model could build a new prompt based on the principle list for the student model. The prompt can then be reused for all instances of that task. This makes \ours significantly more efficient than using a retriever or a teacher model's intervention in each query.

\section{Experiments}

\subsection{Experiments Setup}
\textbf{Datasets.}
We evaluate our approach on eight datasets from diverse domains, including four tasks from Big-bench \citep{srivastava2022beyond}: Tracking Shuffled Objects, Date Understanding, Navigate, and Matrixshapes. The other four datasets are GSM8K \citep{cobbe2021training}, SVAMP \citep{svamp}, CoinFlip and Last Letter Concatenation \citep{kojima2022large, wei2022chain}. We use the 5-object versions of the Tracking Shuffled Objects task in the experiments, which we will refer to as Tracking Shuffled Objects (5). The CoinFlip, Last Letter Concatenation and Tracking Shuffled Objects (5) are regarded as symbolic reasoning tasks, and the remaining datasets are arithmetic reasoning tasks.

For each symbolic reasoning dataset, we sample three questions as the training set and utilize the remaining data as the test set. Notice the CoinFlip and GSM8K is already split into the training set and the test set. We use the original test set and randomly sample 3 questions from the original training set to build our training set. For arithmetic reasoning tasks, we provide practice questions amounting to 25\% of the total number of test questions and 3 questions as the training set. 
Regarding Date Understanding, Navigate and Matrixshapes, we utilize data from \citep{cai2023large}. We divide each dataset into training, validation, and test sets, containing 3, 47, and 200 instances, respectively. Detailed information about each dataset can be found in Appendix \ref{app:datasets}.

\textbf{Experiment settings.}
We experiment with gpt-3.5-turbo-16k and gpt-4, where gpt-4 is the teacher model and gpt-3.5-turbo-16k is the student model. In the following sections, we will denote the two LLMs as GPT3.5 and GPT4, respectively. We use the default temperature of 0.0 for both models. For symbolic reasoning tasks, we utilize CoT as our base prompting method. For arithmetic tasks, we utilize the ``Program of Thought" (PoT) method as our base prompting method. This method involves having the LLMs process natural language questions and create corresponding programs that represent intermediate reasoning steps, with the actual computation of solutions being delegated to a runtime environment, such as a Python interpreter. We avoid using CoT due to the tendency of language models to produce erroneous mathematical operation results when tackling arithmetic tasks \citep{ji2023survey}. Our framework is designed to instruct the student model on problem-solving strategies rather than on performing detailed decimal operations with precision.

\textbf{Baselines.}
Since our framework generates a prompt for the student model for the downstream task, we compare several prompting methods in our experiments. Specifically, we adopt Zero-Shot CoT \citep{kojima2022large}, few-shot CoT \citep{wei2022chain}, and Auto-CoT \citep{zhang2022automatic} as baseline prompting methods. For the few-shot CoT, we use the questions in the training set as exemplars for few-shot prompting. For Auto-CoT, we use Sentence-BERT \citep{reimers2019sentence} to compute a vector representation for each question and then cluster practice questions by \emph{k}-means. Following the original paper, we choose the question closest to the center from each cluster and ask the teacher model to generate reasoning steps with Zero-shot CoT to build final examples for the student model.

\subsection{Symbolic reasoning}

\begin{table}[ht]\small
  \centering
  \caption{Performance on symbolic reasoning and arithmetic tasks, measured in accuracy(\%). The teacher model is GPT4, and the student model is GPT3.5. In symbolic reasoning tasks, we utilize CoT as our base prompting method. In arithmetic reasoning tasks, we utilize PoT as our base prompting method. For \ours, we test the performance with and without error summarization (ES). }
  \label{main_table}
  \scalebox{0.92}{
    \begin{tabular}{c|ccc|ccccc}
      \toprule
      \multicolumn{1}{c}{} & \multicolumn{3}{c}{\textbf{Symbolic reasoning}}                                                                                          & \multicolumn{5}{c}{\textbf{Arithmetic reasoning}}                                    \\ \midrule
      \textbf{Method}  & \textbf{Coin}  & \textbf{Letter} & \textbf{Shuffled (5)} & \textbf{Date} & \textbf{Navigate} & \textbf{GSM8K} & \textbf{Matrix} & \textbf{SVAMP} \\ \midrule
      \multicolumn{1}{c|}{0-shot CoT/PoT}   & 65.7           & 48.5            & 52.4                                                           & 24.0          & 39.0              & 66.7           & 26.5            & 77.6           \\
      \multicolumn{1}{c|}{3-shot CoT/PoT}   & 80.6           & 82.9            & 74.8                                                           & 68.5          & 81.5              & 74.5           & 84.5            & 82.6           \\
      \multicolumn{1}{c|}{Auto CoT/PoT}     & 80.6           & 81.7            & \textbf{75.4}                                                           & 71.0          & 83.0              & 73.9           & 83.5            & 82.8           \\
      \multicolumn{1}{c|}{TPD w/o ES} & \textbf{100.0} & \textbf{89.7}   & 65.8                                                  & 33.5          & 85.0              & 74.7           & 85.0            & 81.2           \\
      \multicolumn{1}{c|}{TPD w/ ES}              & -              & -               & -                                                              & \textbf{76.5} & \textbf{97.5}     & \textbf{75.4}  & \textbf{93.5}   & \textbf{82.9}  \\
      \bottomrule
    \end{tabular}}
\end{table}

Symbolic reasoning entails the use of symbols and their relationships to execute logical operations \citep{maccoll1897symbolic}. It gauges logical reasoning and rule-based decision-making abilities, assessing a language model's proficiency in simulating human-like reasoning. Table~\ref{main_table} presents the comparison results. As anticipated, 3-shot CoT and Auto-CoT surpass zero-shot CoT by providing more examples and guidance to the language model, enabling it to generate higher-quality reasoning processes. Auto CoT achieves comparable performance across all three datasets compared to 3-shot CoT, suggesting that its effectiveness heavily relies on the specific task. Our method \ours outperforms 3-shot CoT on CoinFlip and Last Letter Concatenation, demonstrating the potential effectiveness of problem-solving instructions over merely presenting a sequence of language reasoning steps. However, compared to 3-shot CoT and Auto CoT, problem-solving instructions underperform on Tracking Shuffled Objects (5). This indicates that the quality of problem-solving instructions and examples generated by the teacher model may be inferior to human-written CoT.

We omit the error summarization stage in symbolic reasoning tasks, primarily because errors often stem from a lack of factual knowledge rather than the misapplication of principles. Principles serve as high-level guidelines, applicable to a broad range of scenarios. They do not change with specific situations but offer a consistent approach to problem-solving. In contrast, factual knowledge, which entails specific, verifiable information about the world, is crucial in addressing specific cases. It should be incorporated into the model's parameters during training or sourced from external databases when needed. For instance, consider the Last Letter Concatenation task: the student model adheres to the problem-solving instruction to identify and concatenate the last letters of words to form a new string. The principle guides the process, but errors may occur due to inadequate factual knowledge about the last letter of a particular word within the string. 

\subsection{Arithmetic reasoning}

The results of arithmetic tasks are shown in Table~\ref{main_table}. Zero-shot PoT results in the worst performance in all datasets. The main issue for 0-shot PoT is the LLMs' inability to utilize their pre-trained knowledge to craft solutions for problems without explicit guidance, thereby failing to tap into their potential for coding and logical reasoning with a basic prompt alone. The Auto-PoT approach, which stratifies validation by setting questions into three clusters and selecting the example closest to the center from each cluster to form a 3-shot PoT, aims to introduce diversity into the examples. However, it only marginally improves performance across three datasets and performs poorly on GSM8K and Matrixshapes, which suggests that a mere variety in questions does not inherently lead to distinct reasoning pathways. The original problem-solving instruction derived from the teacher model demonstrates performance comparable with, and occasionally inferior to the 3-shot PoT, since manually created examples may surpass those generated by LLMs in quality.

In comparison, \ours with error summarization outperforms other prompts significantly over Date Understanding, Navigate, and Matrixshape tasks, which indicates that a principle list, distilled from the analysis of errors in practice questions, is highly effective. It guides the teacher model in refining the original problem-solving instruction and choosing the most informative samples for few-shot prompts. However, it only achieves comparable performance with 3-shot PoT on GSM8K and SVAMP, since the errors made by the student model are diverse understanding errors caused by lack of factual knowledge and can not be well categorized in a high-level principle list.

\subsection{How to utilize the principle list} \label{use_principle}

With the principles identified by the teacher model, we explored methods to effectively transfer this knowledge to the student model. One straightforward approach involves directly appending the list of principles to the prompt. Alternatively, we can employ an iterative learning process, where the student model first attempts to answer a question, receives feedback based on the principles, and then revises its initial response. This critique-and-revise strategy has been successfully utilized in various prompting methods (e.g., \citep{chen2023teaching, bai2022constitutional, madaan2023self}). Our proposed method enhances this approach by selecting highly informative examples from the validation set, guided by the principle list. This strategy capitalizes on the emergent capabilities of LLMs, enabling the student model to learn effectively from exemplary instances.

\begin{wraptable}{r}{0.55\textwidth}\small
  \centering
  \caption{Performance of different principle injection methods on arithmetic reasoning tasks, measured in accuracy.}
  \label{inject_exploration}
  \scalebox{0.8}{\begin{tabular}{cccccc}
                   \toprule
                   \textbf{Method}                                                          & \textbf{Date} & \textbf{Navigate} & \textbf{GSM8K} & \textbf{Matrix} & \textbf{SVAMP
                   } \\
                   \midrule
                   \textbf{No principle}                                                    & 68.5          & 81.5              & 74.5           & 84.5            & 82.6           \\
                   \midrule
                   \textbf{\begin{tabular}[c]{@{}c@{}}Injecting\\ Into Prompt\end{tabular}} & 71.5          & 83.5              & 74.8           & 86.5            & 82.6           \\
                   \midrule
                   \textbf{\begin{tabular}[c]{@{}c@{}}Critique\\ + Revise\end{tabular}}     & 61.5          & 74.5              & 71.4           & 79.5            & 74.8           \\
                   \midrule
                   \textbf{\begin{tabular}[c]{@{}c@{}}Examples\\ Selection\end{tabular}}    & \textbf{76.5} & \textbf{97.5}     & \textbf{75.4}  & \textbf{93.5}   & \textbf{82.9}  \\
                   \bottomrule
                 \end{tabular}}
                 \vspace{0.3cm}
             \end{wraptable}

The experimental results presented in Table~\ref{inject_exploration} reveal that simply adding a principles list to the prompt yields only a marginal improvement over the base prompt. This suggests that the student model faces challenges in effectively utilizing high-level and implicit principles expressed in natural language. While retrieval methods typically gather specific factual knowledge and integrate it with the initial prompt, the principles here are more abstract and harder for LLMs to leverage. For instance, a principle might serve as a directional guide for a specific step in the reasoning process. Additionally, LLMs tend to lose information in longer contexts \citep{liu2023lost}, which could further contribute to the limited impact of the principles list.
Surprisingly, the critique-and-revise method resulted in a decrease in performance across all datasets. Our observations indicate that when prompted to provide feedback or critique based on the list of principles, the LLM tends to perceive the original answer as incorrect and significantly overhauls it. This behavior might stem from the RLHF stage, where terms like 'feedback' or 'critique' could trigger the model to question its previous outputs."

Using the principle lists to select examples from the validation set for in-context learning achieves better performance than the other two methods across all datasets. The examples selected based on the principle list contain the most error-prone questions for the student model, thus helping the student model learn from errors effectively. This also mirrors real-world classroom dynamics, where providing students with practical examples often proves more beneficial than solely relying on textbook knowledge \citep{shafto2014rational}.

             \subsection{Ablation study}
             \begin{figure}[ht]
               \centering
               \subfloat[]
               {
                 \label{fig:symbolic_reasoning_num_examples}\includegraphics[width=0.48\textwidth]{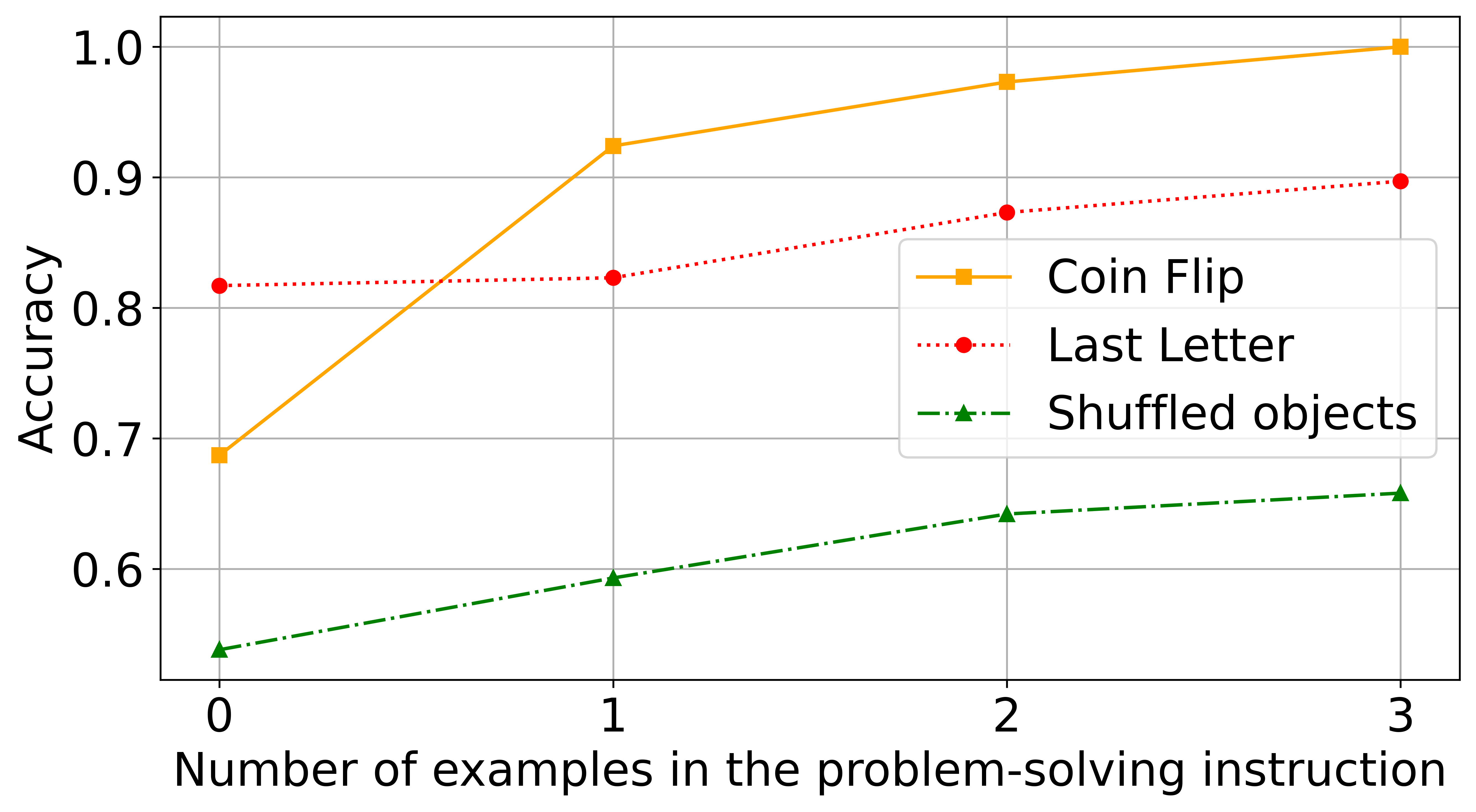}
               }
               \subfloat[]
               {
                 \label{fig:arithmetic_reasoning_num_examples}\includegraphics[width=0.48\textwidth]{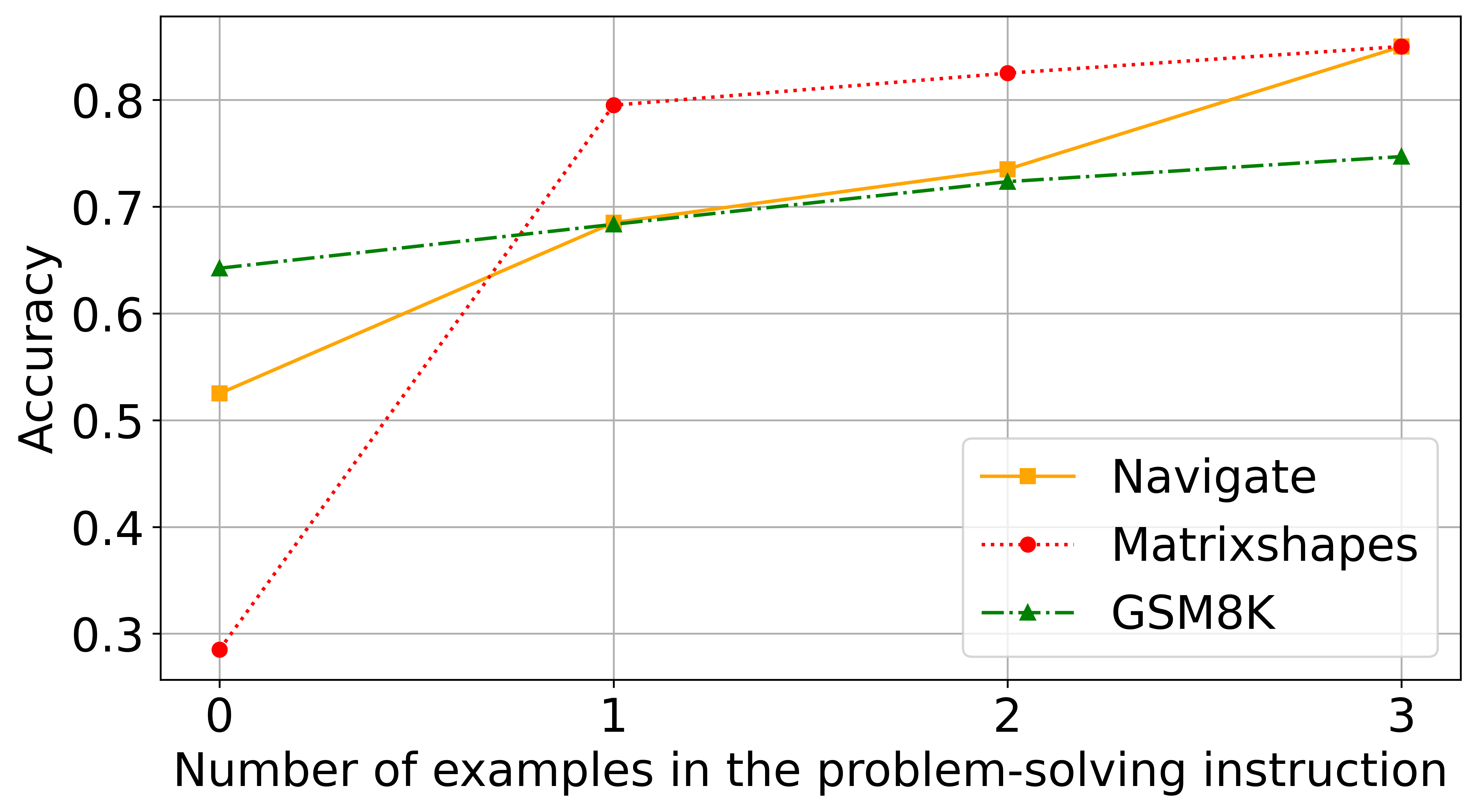}
               }
               \caption{The test accuracy of different numbers of examples in the problem-solving instruction in (a)
                 symbolic reasoning tasks and (b) arithmetic reasoning tasks.}
               \label{fig:num_examples]}
             \end{figure}

             \paragraph{Examples are important in the problem-solving instruction.} We investigate the effectiveness of having the teacher model simply describe a problem-solving method in the instruction without including examples. The results are shown in Fig.~\ref{fig:symbolic_reasoning_num_examples} and Fig.~\ref{fig:arithmetic_reasoning_num_examples}. There is a significant decrease in accuracy for both tasks when examples are omitted from the problem-solving instruction, indicating the importance of incorporating examples in the problem-solving instruction. In more complex tasks like Matrixshapes and Navigate, we observed a notable increase in performance with the inclusion of the first example, while subsequent examples contributed to less. This demonstrates that for complex tasks, the student model struggles to learn from descriptions alone and relies on concrete examples to comprehend the problem-solving instruction. Across all datasets, more examples will help more, but the performance gain will be trivial as the number of examples increases. However, the marginal gains diminish as the number of examples grows, indicating a point of diminishing returns in the value added by additional examples.

\begin{figure}[h]
               \centering
               \subfloat[]
               {
                 \label{fig:exp_in_meta}\includegraphics[width=0.5\textwidth]{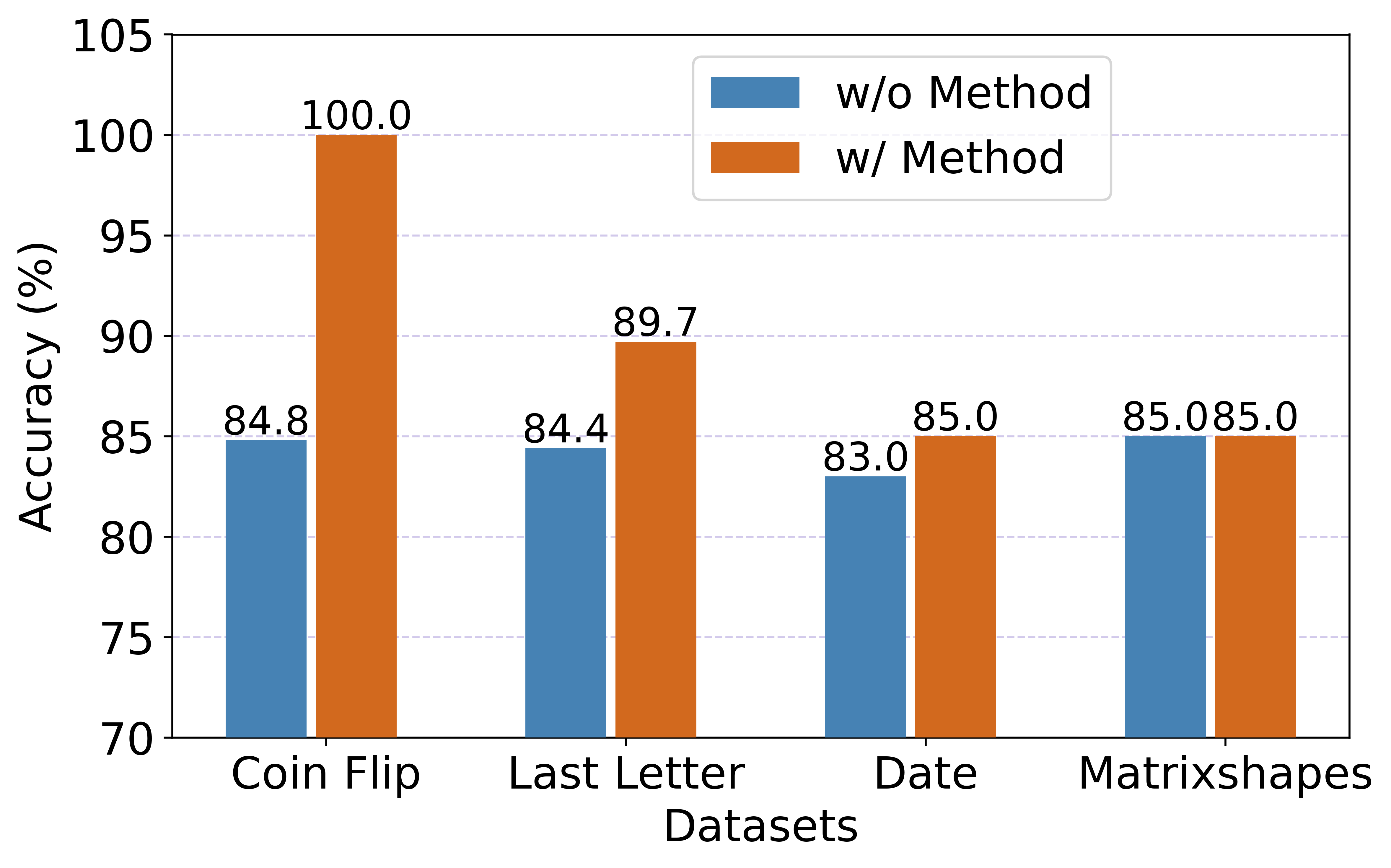}
               }
               \subfloat[]
               {
                 \label{fig:replace_append}\includegraphics[width=0.5\textwidth]{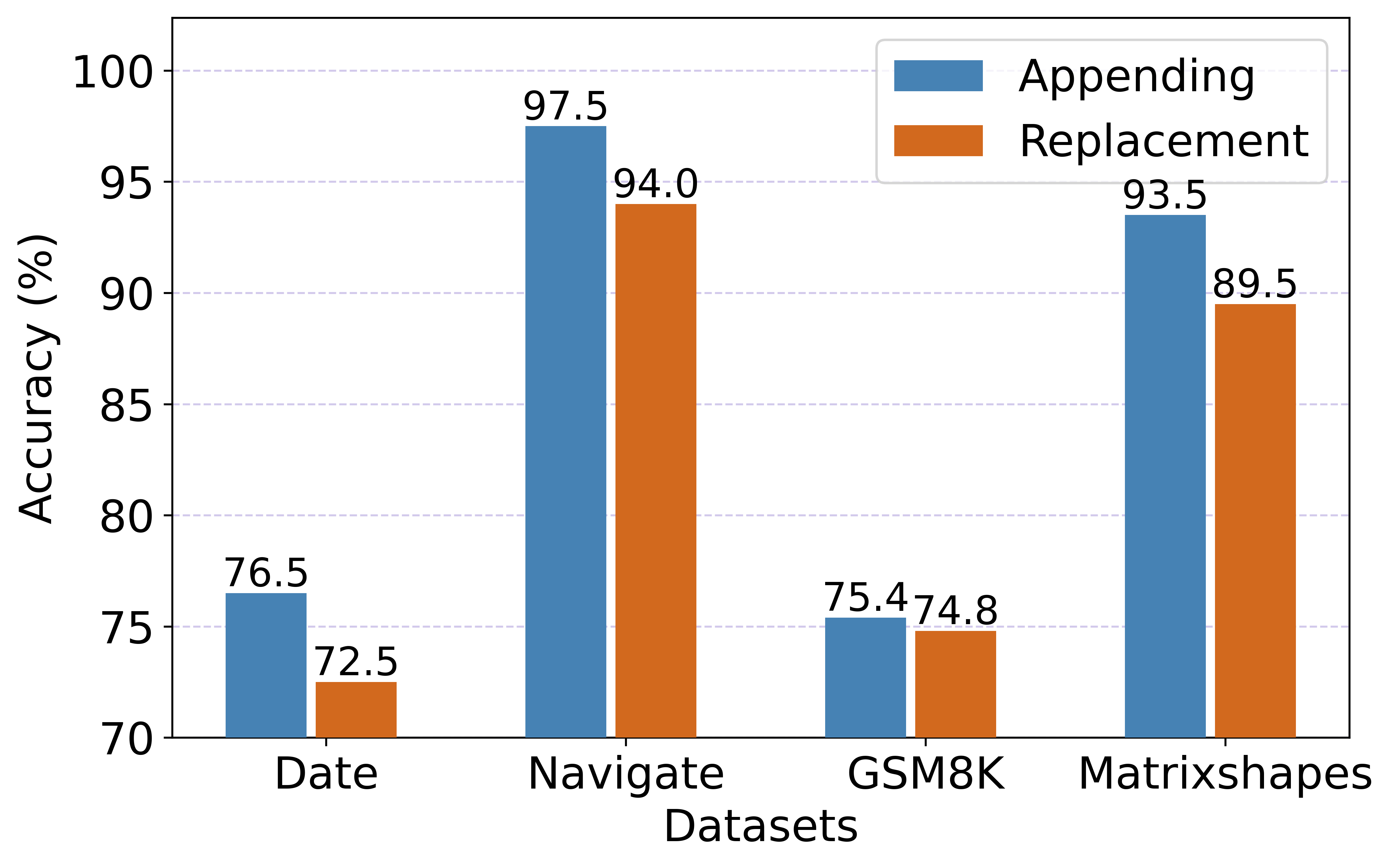}
               }
               \caption{Ablation studies on (a) whether the teacher model needs to provide problem-solving methods in the problem-solving
                 instruction and (b) how to utilize selected examples with the modified problem-solving instruction.}
               \label{fig:ablation}
             \end{figure}

             \paragraph{Methods description in the problem-solving instruction also helps improve the reasoning performance.} Our study also investigates whether the teacher model should offer a high-level reasoning method in problem-solving instruction. As illustrated in Fig. \ref{fig:exp_in_meta}, we observe a notable decline in problem-solving performance without an explicit method description across three datasets. This suggests that while the student model can implicitly learn and mimic reasoning from examples, it remains crucial for the teacher model to explicitly provide a problem-solving method in natural language.

             \begin{wrapfigure}{r}{0.5\textwidth}
  \centering
\includegraphics[width=0.48\textwidth]{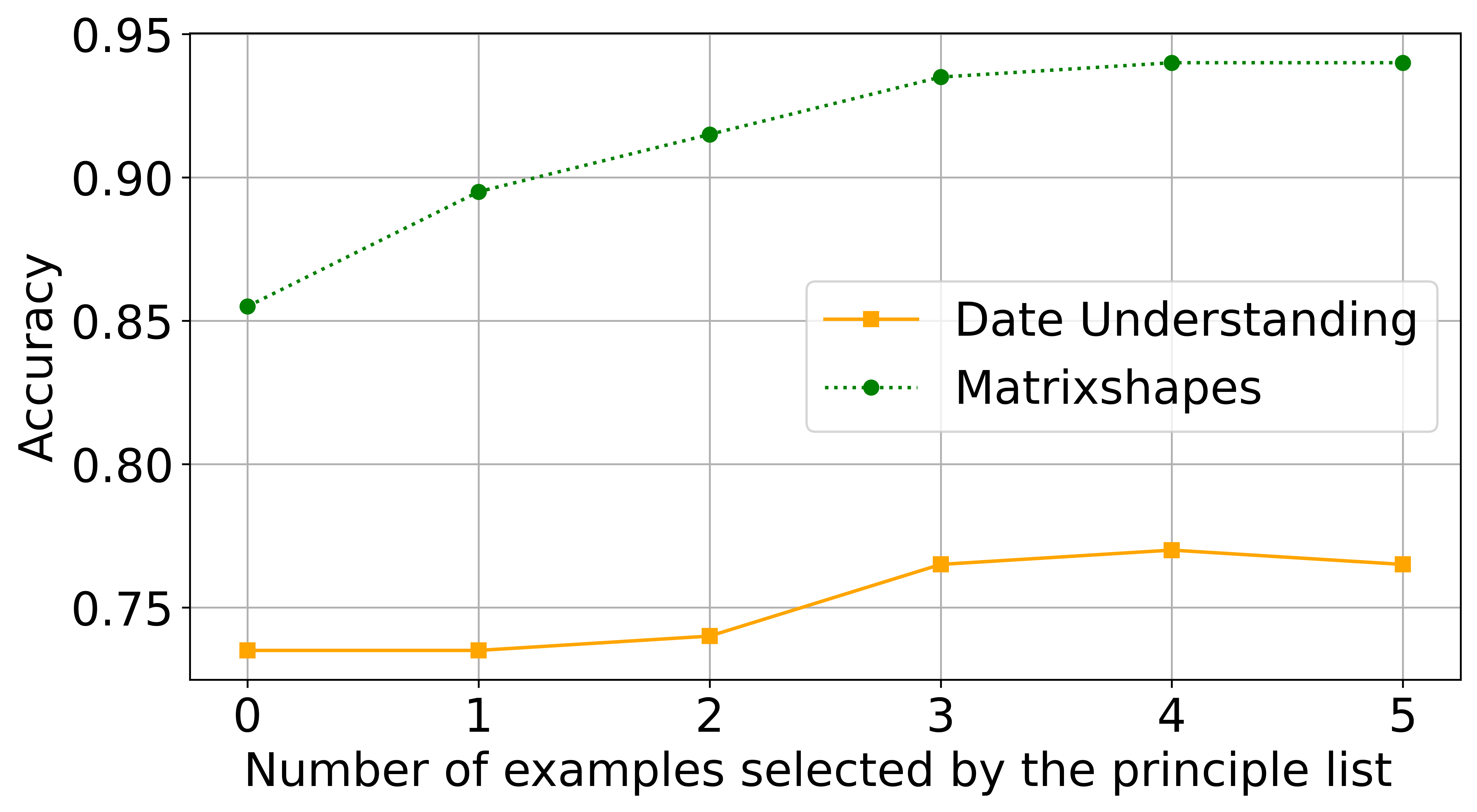}
  \caption{An ablation study on the number of examples selected based on the principle list. 0 example means the prompt only contains the modified problem-solving instruction.
  }
  \label{fig:principle_list_num_examples}
\end{wrapfigure}
             \paragraph{Numbers of examples selected in the principle exploitation stage.} To determine the minimum number of examples required for effective learning, we conducted experiments with different numbers of examples selected based on the violation score. The results are presented in Fig.~\ref{fig:principle_list_num_examples}. Our findings reveal that the initial example provides the most significant performance improvement across both datasets. This suggests that the initial example serves as the most instructive instance for the student model. The minimum number of examples required varies across datasets. For Matrixshapes, 3 examples are sufficient to achieve stable performance. However, for Date Understanding, while 3 examples provide a substantial gain, adding more examples continues to benefit the student model. For simplicity and consistency across all datasets, we employ 3 examples in our experiments.

             \paragraph{Can we replace the original examples in the problem-solving instruction with the newly selected samples?} 
Fig. \ref{fig:principle_list_num_examples} demonstrates the high informativeness of examples selected based on violation scores. This raises an intriguing question: can we directly replace the original examples in the modified problem-solving instruction with these newly chosen ones?
To explore this, Fig. \ref{fig:replace_append} presents a comparison  between two approaches: appending the new selections to the original examples versus completely replacing them. Interestingly, the appending method consistently outperforms the replacement approach across all datasets.
This superior performance can be attributed to the fact that the new selections specifically address the errors encountered when the student model learns from the original examples. Moreover, completely removing the original examples may inadvertently shift the student model's errors, rendering the newly selected examples less informative.

\section{Limitations and Conclusion}

\textbf{Limitations.} One limitation of the \ours is its inability to rectify common sense errors (lack of factual knowledge), as evidenced by its marginal improvements in the GSM8K and SVAMP datasets in Table~\ref{main_table}, despite the teacher model's high accuracy (97\%, \citealp{zhou2023solving}). This limitation arises because the errors made by the student model predominantly relate to factual knowledge, a category not effectively addressed by the generated principles and instructions. High-level principles can resolve only a limited number of such errors. A potential solution for this challenge could involve fine-tuning the LLM to internalize a broader range of common sense knowledge. 
Another constraint of \ours is the method for applying principles. While selecting examples based on validation scores has proven more effective than other strategies, this approach is not particularly efficient. The issue stems from the LLMs' context length limitations. This inefficiency poses an unresolved challenge: how to integrate a long list of principles into LLMs effectively.

\textbf{Conclusion.} In this paper, we present a novel framework named Teaching via Principle Discovery (\ours). This approach empowers teacher models to construct problem-solving instructions and summarize key principles by analyzing example questions and student errors. The identified principles are subsequently employed to refine the problem-solving instructions and select the most informative examples from the validation set to create a tailored instruction prompt. We validate the effectiveness of \ours on symbolic and arithmetic reasoning tasks, observing a marked enhancement in the performance of the student model. Our method introduces an innovative approach, utilizing advanced language models to guide weaker agents in tackling reasoning problems. In the future, we plan to study how to apply \ours to solve complex reasoning tasks, such as serving as a web agent, where the principle list will be much longer. It might be helpful to transform the principle list into a tree structure and then perform some tree search algorithms to find the most appropriate principles and inject them into the student model.


\bibliography{iclr2024_conference}
\bibliographystyle{iclr2024_conference}

\appendix
\section{Datasets}
\label{app:datasets}
The \textbf{Coin Flip} task \citep{wei2022chain} requires the LLM to determine if a coin remains heads up following a series of flips or non-flips. The \textbf{Last Letter Concatenation} task \citep{wei2022chain} requires the LLM to concatenate the last letters of each word in a given name list. Both tasks have out-of-domain test sets, which include more complex examples than those provided in the training exemplars. Specifically, for the Last Letter Concatenation task, the training set contains the lists comprising only two names but the test set consists of the lists with three and four names. A similar setting is adopted for the Coin Flip task.

\textbf{Tracking shuffled objects} \citep{srivastava2022beyond} requires the LLM to determine the ultimate state of a system based on its initial condition and a series of subsequent changes. In each instance of this task, a collection of objects is initially associated with individual owners. These objects are then exchanged in a sequence of swappings. The model's objective is to determine the final owner of each object.

\textbf{Date Understanding} \citep{srivastava2022beyond} asks the LLM to infer the date from the given context. The context is one or two sentences with date information. 

\textbf{Navigate} \citep{srivastava2022beyond} requires the LLM to infer the agent's position after several movements. Specifically, given a series of navigation steps, the LLM needs to determine whether or not an agent would end up back at the starting point.

\textbf{GSM8K} \citep{cobbe2021training} is a collection of grade-level mathematics problems by human authors. These problems require 2 to 8 steps for resolution, primarily requiring a series of fundamental computations employing basic arithmetic operations (addition, subtraction, multiplication, and division) to arrive at the solution.

\textbf{Matrixshapes} \citep{srivastava2022beyond} requires the LLM to predict the shape of the result of a chain of matrix manipulations, given the inputs' shapes. 

\textbf{SVAMP} \citep{svamp} is a challenge set for elementary-level Math Word Problems. It consists of questions that test a model across question sensitivity, reasoning ability, and invariance to structural alterations. 

Table~\ref{ds_example} shows the example questions for each dataset.

\begin{table}[!htbp]
\centering
\caption{Examples of questions in each dataset.}
\label{ds_example}
\scalebox{0.94}{
    \begin{tabularx}{\textwidth}{|X|}
        \toprule
        \multicolumn{1}{|c|}{\textbf{Coin Flip}} \\ 
        \midrule
        \textbf{Q}: A coin is heads up. Murraylee does not flip the coin. Meilich flips the coin. Is the coin still heads up? \\
        \textbf{A}: no \\
        \midrule
        \multicolumn{1}{|c|}{\textbf{Last Letter Concatenation}} \\
        \midrule
        \textbf{Q}: Take the last letters of each word in "Maritza Nana Loretta Eric" and concatenate them. \\
        \textbf{A}: "aaac" \\
        \midrule
        \multicolumn{1}{|c|}{\textbf{Tracking shuffled objects}} \\
        \midrule
        \textbf{Q}: Alice, Bob, Claire, Dave, and Eve are dancers at a square dance. At the start of a song, they each have a partner: Alice is dancing with Patrick, Bob is dancing with Sam, Claire is dancing with Jamie, Dave is dancing with Lola, and Eve is dancing with Melissa. Throughout the song, the dancers often trade partners. First, Dave and Eve switch partners. Then, Dave and Alice switch partners. Then, Eve and Alice switch partners. Then, Claire and Bob switch partners. Finally, Dave and Alice switch partners. At the end of the dance, Alice is dancing with \\
        Options: \\
        (A) Patrick \\
        (B) Sam \\
        (C) Jamie \\
        (D) Lola \\
        (E) Melissa \\
        \textbf{A}: (A) \\
        \midrule
        \multicolumn{1}{|c|}{\textbf{Date Understanding}} \\
        \midrule
        \textbf{Q}: Jane scheduled 3 appointments with 5 people for tomorrow (Tue, 7/9/1972). What is the date a month ago in MM/DD/YYYY? \\
        Options: \\
        (A) 06/08/2059 \\
        (B) 06/22/1972 \\
        (C) 12/08/1971 \\
        (D) 06/08/2034 \\
        (E) 06/08/1972 \\
        (F) 06/07/1972 \\
        \textbf{A}: (E) \\
        \midrule
        \multicolumn{1}{|c|}{\textbf{Navigate}} \\
        \midrule
        \textbf{Q}: If you follow these instructions, do you return to the starting point? Always face forward. Take 1 step backward. Take 9 steps left. Take 2 steps backward. Take 6 steps forward. Take 4 steps forward. Take 4 steps backward. Take 3 steps right. \\
        Options: \\
        - Yes \\
        - No \\
        \textbf{A}: No \\
        \midrule
        \multicolumn{1}{|c|}{\textbf{GSM8K}} \\
        \midrule
        \textbf{Q}: Megan is an actress. She was the lead actress in 80\% of her work. In total, Megan participated in 100 plays. How many times was Megan not the lead actress? \\
        \textbf{A}: 20.0 \\
        \midrule
        \multicolumn{1}{|c|}{\textbf{SVAMP}} \\
        \midrule
        \textbf{Q}: In a school there are 308 girls and 318 boys. There are also 36 teachers. How many pupils are there in that school? \\
        \textbf{A}: 626.0 \\
        \midrule
        \multicolumn{1}{|c|}{\textbf{Matrixships}} \\
        \midrule
        \textbf{Q}: Keep track of matrix shapes through various transformations. Transpose a matrix of shape (2,3,2). Transpose the result. Compute the Hadamard product of the result with a matrix of shape (2,3,2). Compute the Hadamard product of the result with a matrix of shape (2,3,2). Sum the result over the second axis. \\
        \textbf{A}: (2,2) \\
        \bottomrule
    \end{tabularx}
}%
\end{table}%

\section{Experiment Setup}

\subsection{GPT versions}
In the experiments, we utilize two GPT models: gpt-3.5-turbo-16k and gpt-4. The gpt-3.5-16k refers to the ``gpt-3.5-turbo-16k'' model and in the OpenAI API model with checkpoint version 2023-06-13-preview webpage.\footnote{\href{https://platform.openai.com/docs/models}{https://platform.openai.com/docs/models}}
GPT-4 refers to the ``gpt-4'' model with checkpoint version 2023-07-01-preview. All mentioned checkpoints are hosted on Microsoft Azure,\footnote{\ *.openai.azure.com} designated for our group's specific usage.
The model temperature is $0$ in all cases.

\subsection{Principle Lists}

Table~\ref{principle_example_1} and Table~\ref{principle_example_2} show the principle list the teacher model finds in the error summarization stage. These principles provide high-level guidance applicable across various scenarios. However, the principle lists for GSM8K and SVAMP are less clear compared to others, since the errors in these tasks primarily relate to factual knowledge.

\begin{table}[!htbp]
\caption{Examples of principle list.}
\label{principle_example_1}
\begin{tabularx}{\textwidth}{|X|}
\toprule
\multicolumn{1}{|c|}{\textbf{Svamp}}                                                             \\ \midrule
1. Incorrect Mathematical Operations: Ensure that the correct mathematical operations are used to solve the problem. Misinterpretation of the problem statement often leads to incorrect operations. \\
2. Misinterpretation of Problem Requirements: Understand the core requirement of the question. \\
3. Logical Errors in Variable Initialization: Be cautious when initializing variables. Make sure the initial values correctly represent the situation described in the question. \\
4. Misapplication of Variables in Calculation: Ensure that the variables are applied correctly in the calculation formula. \\
5. Understanding the Context of the Question: Contextual understanding is crucial. In the question about pots, flowers, and sticks, it's important to realize that the total number of flowers and sticks is a cumulative count across all pots, requiring multiplication of the per pot count by the total number of pots.\\ \midrule
\multicolumn{1}{|c|}{\textbf{Date Understanding}}                                                                                                                                                                         \\ \midrule
1. Understanding Date Arithmetic and the 'datetime' Module: Several examples demonstrate a misunderstanding of how the 'datetime' module works, especially in terms of adding or subtracting days, months, and years. The 'timedelta' function in Python doesn't support months or years directly, so programmers need to account for this limitation when performing date arithmetic. Instead of using 'timedelta' for date manipulations, use 'dateutil.relativedelta'. This module provides more flexibility, especially for operations involving months and years.  \\
2. Accurate Date Initialization: Initialize dates correctly. In several examples, the initial date is set without considering the context of the problem. Ensure that the starting point of the calculation aligns with the scenario's requirements. \\
3. Logical Consistency in Calculations: Maintain logical consistency in calculations. If the problem states a historical or future date, ensure that the calculations reflect this timeline accurately. Avoid mixing current dates ('datetime.now()') with historical or future scenarios unless it's relevant.\\
4. Validating Against Given Options: When comparing calculated dates against multiple-choice options, ensure that the options are correctly formatted and compared. It's essential to format the calculated date in the same format as the options for a valid comparison.\\ \midrule
\multicolumn{1}{|c|}{\textbf{Navigate}}                                                                         \\ \midrule
1. Understand the Problem: Recognize that the problem requires tracking movements in two dimensions (horizontal and vertical). Understand that movements are influenced by the current direction the subject is facing. Identify the need to interpret turns as changes in direction, not just movement. \\
2. Initialize Variables: Define variables for horizontal and vertical movements, initializing them to zero. Introduce a variable for current direction, initializing it to the starting orientation (e.g., "north"). \\
3. Extract and Interpret Instructions: Parse the given question to extract movement and turning instructions. \\
4. Use control structures (if-else, loops) to handle each instruction:
For movement instructions (forward, backward), update the horizontal or vertical position based on the current direction. For turning instructions (right, left, around), update the current direction appropriately.
5. Handle Directional Changes:
Implement logic to correctly modify the direction state when turning. For example, turning right from north means facing east. \\
6. Calculate Final Position:
After processing all instructions, compare the final horizontal and vertical positions with the initial position (0,0). \\
7. Return Result:
Return the appropriate option ("Yes" or "No") based on whether the final position matches the starting point. \\
8. Implement Directional Logic:
Develop a mechanism to translate turning instructions into directional changes, affecting subsequent movement calculations. \\
9. Consider Special Cases: Account for any special or compound instructions that may require separate handling, ensuring all scenarios are covered.                                    \\ \midrule

\end{tabularx}%
\end{table}%

\begin{table}[!htbp]
\caption{Examples of principle list.}
\label{principle_example_2}
\begin{tabularx}{\textwidth}{|X|}
\toprule
\multicolumn{1}{|c|}{\textbf{GSM8K}} \\ 
\midrule
1. Incorrect Variable Assignment and Utilization: Ensure variable assignments accurately reflect the values they are supposed to represent based on the question's context. \\
2. Misinterpretation of Quantity Relationships: Accurately understand and interpret the relationships between different quantities as described in the problem statement. \\
3. Incorrect Mathematical Formulas: Ensure the mathematical formulas used align with the logical requirements of the problem. \\
4. Misinterpretation of Variable Values: Ensure variables are interpreted and utilized accurately based on the problem’s context. \\
5. Omission of Critical Information: Incorporate all provided information and ensure no critical details are omitted in the solution. \\
6. Incorrect Arithmetic Operations: Ensure arithmetic operations are logically sound and mathematically correct. \\ 
\midrule
\multicolumn{1}{|c|}{\textbf{Matrixships}} \\ 
\midrule
1. Matrix Multiplication: Use np.matmul for matrix multiplication. \\
2. Hadamard Product: Use * (asterisk) for element-wise multiplication (Hadamard product). \\
3. Transposition: Utilize np.transpose for matrix transposition. Do not specify the axes parameter when using np.transpose. \\
4. Resultant Matrix Shape: Always return the shape of the resulting matrix after an operation. \\
\midrule
\end{tabularx}%
\end{table}%

\subsection{Problem-solving instruction examples}
We provide an example of problem-solving instruction for a better understanding. The problem-solving instruction consists of a problem-solving method and several examples showing how to use the method. The questions in the examples are training questions.
\begin{lstlisting}[caption=An example of problem-solving instructions provided by the teacher model.]
###Problem-solving instruction for CoinFlip task

Method to solve coin filp problems:
1. Start by reading the initial state of the coin (heads up or tails up).
2. Examine the actions of each person mentioned in the question in the order they are mentioned. If a person flips the coin, the state of the coin will change. If it was heads up, it will now be tails up, and vice versa.
If a person does NOT flip the coin, the state remains unchanged. At the end of the actions, state the final position of the coin as 'heads up' or 'tails up'.
3. Answer the question based on the final state of the coin compared to the state asked in the question.

Examples:

Question: A coin is heads up. sager does not flip the coin. zyheir flips the coin. Is the coin still heads up?
Explanation:
The coin starts as heads up.
Sager does not flip the coin, so it remains heads up.
Zyheir flips the coin, changing its state. Now, it's tails up.
The final position of the coin is tails up.

Answer: No.

Question: A coin is heads up. mailey does not flip the coin. maurisa does not flip the coin. Is the coin still heads up?
Explanation:
The coin starts as heads up.
Mailey does not flip the coin, so it remains heads up.
Maurisa also does not flip the coin, so it stays heads up.
The final position of the coin is heads up.

Answer: Yes.

Question: A coin is heads up. murraylee does not flip the coin. meilich flips the coin. Is the coin still heads up?

Explanation:
The coin starts as heads up.
Murraylee does not flip the coin, so it remains heads up.
Meilich flips the coin, changing its state. Now, it's tails up.
The final position of the coin is tails up.
Answer: No.
\end{lstlisting}

\subsection{Overall student prompt examples}
We provide an example of the overall teacher model's instruction for a better understanding. The overall instruction prompt consists of the modified problem-solving instruction and several newly selected examples from the validation set. Examples 1-3 are original examples in the problem-solving instruction (from the training set). Examples 4-6 are newly selected examples from the validation set.
\begin{lstlisting}[caption=An examples of overall prompt provided by the teacher model.]
###Overall prompts for 'Navigate' task

To write Python functions to answer questions about returning to the starting point after following a set of movement instructions, follow these steps:

Understand the Problem:
Recognize that the problem requires tracking movements in two dimensions (horizontal and vertical).
Understand that movements are influenced by the current direction the subject is facing.
Identify the need to interpret turns as changes in direction, not just movement.

Design the Function:

    Initialize Variables:
    Define variables for horizontal and vertical movements, initializing them to zero.
    Introduce a variable for current direction, initializing it to the starting orientation (e.g., "north").
    Extract and Interpret Instructions:
    Parse the given question to extract movement and turning instructions.
    Use control structures (if-else, loops) to handle each instruction:
    For movement instructions (forward, backward), update the horizontal or vertical position based on the current direction.
    For turning instructions (right, left, around), update the current direction appropriately.
    Handle Directional Changes:
    Implement logic to correctly modify the direction state when turning. For example, turning right from north means facing east.
    Calculate Final Position:
    After processing all instructions, compare the final horizontal and vertical positions with the initial position (0,0).
    Return Result:
    Return the appropriate option ("Yes" or "No") based on whether the final position matches the starting point.

Implement Directional Logic:
Develop a mechanism to translate turning instructions into directional changes, affecting subsequent movement calculations.

Test the Function:
Write test cases to execute the function with different sets of instructions.
Store and verify the result in a variable answer to ensure the function behaves as expected.

Here are some examples:

Question: 'If you follow these instructions, do you return to the starting point? Always face forward. Take 1 step backward. Take 9 steps left. Take 2 steps backward. Take 6 steps forward. Take 4 steps forward. Take 4 steps backward. Take 3 steps right. Options: - Yes - No'

Answer:
```python
def do_you_return_to_starting_point():
    # Initial position and direction
    horizontal, vertical = 0, 0
    direction = "north"

    # Movement instructions
    instructions = [
        ("backward", 1),
        ("left", 9),
        ("backward", 2),
        ("forward", 6),
        ("forward", 4),
        ("backward", 4),
        ("right", 3)
    ]

    # Process each instruction
    for action, steps in instructions:
        if action == "forward":
            if direction == "north":
                vertical += steps
        elif action == "backward":
            if direction == "north":
                vertical -= steps
        elif action == "left":
            if direction == "north":
                horizontal -= steps
        elif action == "right":
            if direction == "north":
                horizontal += steps

    # Check if returned to the starting point
    return "Yes" if horizontal == 0 and vertical == 0 else "No"

# Execute the function
answer = do_you_return_to_starting_point()
```
By using the python function above, you can get the answer of the question.
###END

Question: 'If you follow these instructions, do you return to the starting point? Always face forward. Take 10 steps left. Take 10 steps forward. Take 7 steps forward. Take 2 steps forward. Options: - Yes - No'

Answer:
```python
def do_you_return_to_starting_point():
    # Initial position and direction
    horizontal, vertical = 0, 0
    direction = "north"

    # Movement instructions
    instructions = [
        ("left", 10),
        ("forward", 10),
        ("forward", 7),
        ("forward", 2)
    ]

    # Process each instruction
    for action, steps in instructions:
        if action == "forward":
            if direction == "north":
                vertical += steps
        elif action == "left":
            if direction == "north":
                horizontal -= steps

    # Check if returned to the starting point
    return "Yes" if horizontal == 0 and vertical == 0 else "No"

# Execute the function
answer = do_you_return_to_starting_point()
```
By using the python function above, you can get the answer of the question.
###END

Question: 'If you follow these instructions, do you return to the starting point? Always face forward. Take 1 step right. Take 3 steps left. Take 2 steps right. Options: - Yes - No'

Answer:
```python
def do_you_return_to_starting_point():
    # Initial position and direction
    horizontal, vertical = 0, 0
    direction = "north"

    # Movement instructions
    instructions = [
        ("right", 1),
        ("left", 3),
        ("right", 2)
    ]

    # Process each instruction
    for action, steps in instructions:
        if action == "left":
            if direction == "north":
                horizontal -= steps
        elif action == "right":
            if direction == "north":
                horizontal += steps

    # Check if returned to the starting point
    return "Yes" if horizontal == 0 and vertical == 0 else "No"

# Execute the function
answer = do_you_return_to_starting_point()
```
By using the python function above, you can get the answer of the question.
###END

Question: 'If you follow these instructions, do you return to the starting point? Take 3 steps. Turn around. Take 3 steps. Turn right.\nOptions:\n- Yes\n- No'

Answer:
```python
def do_you_return_to_starting_point():
    # Initial position and direction
    horizontal, vertical = 0, 0
    direction = "north"

    # Movement instructions
    instructions = [
        ("forward", 3),
        ("turn", "around"),
        ("forward", 3),
        ("turn", "right")
    ]

    # Process each instruction
    for action, value in instructions:
        if action == "forward":
            if direction == "north":
                vertical += value
            elif direction == "south":
                vertical -= value
            elif direction == "east":
                horizontal += value
            elif direction == "west":
                horizontal -= value
        elif action == "turn":
            if value == "around":
                direction = "south" if direction == "north" else "north"
            elif value == "right":
                if direction == "north":
                    direction = "east"
                elif direction == "east":
                    direction = "south"
                elif direction == "south":
                    direction = "west"
                elif direction == "west":
                    direction = "north"

    # Check if returned to the starting point
    return "Yes" if horizontal == 0 and vertical == 0 else "No"

# Execute the function
answer = do_you_return_to_starting_point()
```
By using the python function above, you can get the answer of the question.
###END

Question: 'If you follow these instructions, do you return to the starting point? Take 3 steps. Turn around. Take 5 steps. Turn right. Turn right. Take 1 step. Take 1 step.\nOptions:\n- Yes\n- No'

Answer:
```python
def do_you_return_to_starting_point():
    # Initial position and direction
    horizontal, vertical = 0, 0
    direction = "north"

    # Movement instructions
    instructions = [
        ("forward", 3),
        ("turn", "around"),
        ("forward", 5),
        ("turn", "right"),
        ("turn", "right"),
        ("forward", 1),
        ("forward", 1)
    ]

    # Process each instruction
    for action, value in instructions:
        if action == "forward":
            if direction == "north":
                vertical += value
            elif direction == "south":
                vertical -= value
            elif direction == "east":
                horizontal += value
            elif direction == "west":
                horizontal -= value
        elif action == "turn":
            if value == "around":
                direction = "south" if direction == "north" else "north"
            elif value == "right":
                if direction == "north":
                    direction = "east"
                elif direction == "east":
                    direction = "south"
                elif direction == "south":
                    direction = "west"
                elif direction == "west":
                    direction = "north"

    # Check if returned to the starting point
    return "Yes" if horizontal == 0 and vertical == 0 else "No"

# Execute the function
answer = do_you_return_to_starting_point()
```
By using the python function above, you can get the answer of the question.
###END

Question: 'If you follow these instructions, do you return to the starting point? Take 1 step. Take 5 steps. Turn around. Turn around. Turn around. Take 6 steps.\nOptions:\n- Yes\n- No'

Answer:
```python
def do_you_return_to_starting_point():
    # Initial position and direction
    horizontal, vertical = 0, 0
    direction = "north"

    # Movement instructions
    instructions = [
        ("forward", 1),
        ("forward", 5),
        ("turn", "around"),
        ("turn", "around"),
        ("turn", "around"),
        ("forward", 6)
    ]

    # Process each instruction
    for action, value in instructions:
        if action == "forward":
            if direction == "north":
                vertical += value
            elif direction == "south":
                vertical -= value
            elif direction == "east":
                horizontal += value
            elif direction == "west":
                horizontal -= value
        elif action == "turn":
            if value == "around":
                direction = "south" if direction == "north" else "north"

    # Check if returned to the starting point
    return "Yes" if horizontal == 0 and vertical == 0 else "No"

# Execute the function
answer = do_you_return_to_starting_point()
```
By using the python function above, you can get the answer of the question.
###END



\end{lstlisting}

\end{document}